\documentclass[twoside]{article}

\usepackage[accepted]{aistats2022}

\usepackage[utf8]{inputenc} 
\usepackage[T1]{fontenc}    
\usepackage[table,xcdraw]{xcolor}         
\usepackage{hyperref}       
\hypersetup{colorlinks=false,
linkbordercolor=red,linkcolor=false,pdfborderstyle={/S/U/W 0}} 
\usepackage{url}            
\usepackage{booktabs}       
\usepackage{amsfonts}       
\usepackage{nicefrac}       
\usepackage{microtype}      
\usepackage{amsmath,graphicx,amssymb}
\usepackage{amsthm}
\usepackage{wrapfig}

\usepackage{caption}
\usepackage{subcaption}
\usepackage{tikz}
\usetikzlibrary{tikzmark,shapes.geometric, arrows,calc,hobby}

\newtheorem{theorem}{Theorem}

\newtheorem{definition}{Definition}
\newtheorem{remark}{Remark}
\newtheorem{prop}{Proposition}

\def\R{{\mathbb{R}}}
\def\C{{\mathbb{C}}}

\newcommand{\iu}{{i\mkern1mu}}

\DeclareMathOperator{\diag}{diag}
\DeclareMathOperator{\Tr}{Tr}
\begin{document}

%
\runningtitle{Nonstationary multi-output Gaussian processes via harmonizable spectral mixtures}

%

\twocolumn[

\aistatstitle{Nonstationary multi-output Gaussian processes\\ via harmonizable spectral mixtures}

\aistatsauthor{ Matías Altamirano \And Felipe Tobar}

\aistatsaddress{ Departament of Mathematical Engineering\\
  Universidad de Chile\\
  \texttt{maltamirano@dim.uchile.cl} \And  Initiative for Data \& Artificial Intelligence\\
  Universidad de Chile\\
  \texttt{ftobar@uchile.cl} } ]

\begin{abstract}
  Kernel design for Multi-output Gaussian Processes (MOGP) has received increased attention recently. In particular, the Multi-Output Spectral Mixture kernel (MOSM) approach has been praised as a general model in the sense that it extends other approaches such as Linear Model of Corregionalization, Intrinsic Corregionalization Model and Cross-Spectral Mixture. MOSM relies on Cramér’s theorem to parametrise the power spectral densities (PSD) as a Gaussian mixture, thus, having a structural restriction: by assuming the existence of a PSD, the method is only suited for multi-output stationary applications. We develop a nonstationary extension of MOSM by proposing the family of harmonizable kernels for MOGPs, a class of kernels that contains both stationary and a vast majority of non-stationary processes. A main contribution of the proposed harmonizable kernels is that they automatically identify a possible nonstationary behaviour meaning that practitioners do not need to choose between stationary or non-stationary kernels. The proposed method is first validated on synthetic data with the purpose of illustrating the key properties of our approach, and then compared to existing MOGP methods on two real-world settings from finance and electroencephalography.
\end{abstract}

\section{INTRODUCTION}
\label{introduction}


Gaussian Processes (GPs) provide a flexible and powerful non-parametric framework for Bayesian inference on time series, and thus are considered in many areas of application \cite{williams2006gaussian}. The main aspect on the design of the GP is the choice of the covariance function (also called kernel), which encapsulates all the properties of the process such as smoothness, periodicity and stationarity to name a few. The extension of the GP to multiple outputs is known as multi-output Gaussian processes (MOGP) \cite{alvarez2012kernels}, which models the outputs to be jointly Gaussian an thus is able to share information across outputs, potentially improving the estimation. As in the single-output case, designing kernels that successfully
model auto- and cross-covariances between channels is a core challenge in MOGPs.


There have been several approaches to design valid MOGP kernels \cite{alvarez2008sparse,alvarez2012kernels,goovaerts1997geostatistics}, and a number of them  are based on linear combinations of latent-factor independent Gaussian processes. These approaches, though they work in practice, avoid the direct parametrisation of multioutput covariances and may result in constrained covariances, especially from a spectral analysis perspective. Recently, \cite{parra2017spectral} proposed the multi-output spectral mixture (MOSM) which directly designs the kernel in the spectral domain, using the multivariate version of Bochner’s theorem \cite{bochner1959lectures}, namely Cramér’s Theorem \cite{cramer1940theory}. 


The MOSM kernel provides a unified perspective of existing MOGP kernels in the literature, however, its principal limitation is that it is restricted to stationary data, i.e., $k(x,x')=k(x-x')$, thus it encodes an identical similarity notion across the input space. This assumption of stationarity is unsuitable for several real world settings, such as vibratory signals \cite{kim2005damage,zhang2003novel}, free-drifting oceanic instruments \cite{lilly2011analysis}, various neuroscience applications \cite{cranstoun2002time,ombao2001automatic}, and econometrics \cite{joyeux1980harmonizable}. Therefore, a flexible non-stationary multioutput kernel becomes necessary and in particular a non-stationary version of the MOSM kernel. 


In this article, we propose an expressive and flexible family of MOGP kernels to model non-stationary processes as a natural extension of the MOSM. The proposed kernel relies on the concept of harmonizability, a term introduced in \cite{loeve1978probability} which generalizes the Fourier spectral representations to non-stationary processes. The harmonizable processes have been widely studied and developed in the statistics community, but there has been a lack of attentiveness among machine learning researchers.


The rest of the paper is organized as follows. We revisit the required concepts supporting our proposal from the GP literature and the concept of harmonizable processes in Section \ref{background}. Section \ref{proposal} presents the proposed MOGP kernel and Section \ref{sec:previous_work} compares it to previous approaches in the literature. We explain practical considerations of the proposed model in Section \ref{sec:practical_cons} and Section \ref{experiments} validates it on synthetic and real-world data. Lastly, we discuss our results and summarise our contribution and future work in Section \ref{conclusion}.

\section{BACKGROUND}
\label{background}
In this section we briefly introduce Gaussian processes, multi-output Gaussian processes, spectral mixture kernels and harmonizable processes; these support the development of the proposed kernel in Section \ref{proposal}.

\subsection{Gaussian processes}

The GP is a Bayesian nonparametric generative model for functions $f:\mathbb{R}^{D}\to\mathbb{R}$. The GP is the infinite-dimensional extension of the multivariate normal (MVN) distribution, meaning that it can model second-order relationships among an infinite number of random variables. With mean function $m$ and covariance function $k$, a GP denoted by 
$$f\sim \mathcal{GP}(m,k),$$
has the property that any collection of inputs $\{x_1,\dots,x_N\}\subset\mathbb{R}^{D}$, the output $[f(x_1),\dots,f(x_N)]\in\mathbb{R}^{N}$ is distributed according to an MVN of mean $m(\textbf{x})$ and covariance $k(\textbf{x},\textbf{x})$, where $\textbf{x}=[x_1,\dots,x_N]$. 

The design of the GP, involves choosing the kernel function, which determines key properties in the draws from the GP such as differentiability, periodicity, long-range correlation or stationarity. Furthermore, we say that a kernel is stationary if it can be written as $k(x,x') = k(x-x')$; a GP is said to be stationary if its covariance is stationary. The central result on the characterization of stationary kernels is Bochner's theorem: 

\begin{theorem}[\cite{bochner1959lectures}]
    A complex-valued function $k$ on $\R^d$ is the covariance function of
    a weakly-stationary mean-square-continuous stochastic process on $\R^d$ if and only if it admits the following representation
    \begin{equation}
        \label{Bochner}
        k(x,x') = \int_{\R^d}e^{i\omega^\top(x-x')}S(\omega)d\omega,
    \end{equation}
    
    where $S(\omega)$ is a non-negative bounded function on $\R^d$, called the power spectral density, and $i$ denotes the imaginary unit.
\end{theorem}

This theorem defines a duality between time and frequency, which allows us to construct new kernels via their parametrization in the (Fourier) frequency domain. Perhaps the main approach to kernel design in this fashion is the spectral mixture (SM) kernel proposed by \cite{wilson2013gaussian}.

\subsection{Multi-output Gaussian processes}

The extension of the GP framework to handle multiple output is referred to as Multi-Output Gaussian Process (MOGP) \cite{bonilla2007multi}, which consists in modeling all the outputs as jointly Gaussian where the covariance and cross covariance are ruled by a multi-output kernel. In more detail, if we have $M$ output latent function $\{f_i\}_{i=1}^{M}$, the element of $(i,j)$ of the covariance kernel $\mathcal{K}$ corresponds to the covariance between outputs $f_i$ and $f_j$, following the next notation:
\begin{equation}
    \text{cov}[f_i(x),f_j(x')] = k_{ij}(x,x')=[\mathcal{K}(x,x')]_{ij}.
\end{equation}

A kernel function must be symmetric and positive-definite in order to be a valid covariance function, and similar to the single channel case, a multivariate kernel $\mathcal{K}$ is stationary if $\mathcal{K}(x,x')=\mathcal{K}(x-x')$. The design of valid and expressive multi-output kernels is quite challenging because we need to jointly choose functions that model the covariance of each channel and functions that model the cross-covariance between channels \cite{alvarez2012kernels}. Several approaches have been proposed to overcome this difficulty, see, e.g.,  \cite{alvarez2008sparse,goovaerts1997geostatistics,seeger2005semiparametric,ulrich2015gp}. However, mostly all of them are based on the idea of model the cross covariances as a linear combination of the covariance of each channel. In the next subsection we review one of the most prominent extensions of the existing methods in expressiveness and interpretation, the multi-output spectral mixture kernel.

\subsection{Multi-output spectral mixture kernel}

Relying on Cramér's theorem \cite{cramer1940theory}, the multivariate version of the Bochner theorem, \cite{parra2017spectral} provided a new approach to design multivariate covariance functions which allows full parametric interpretation of the relationship across channels, in addition to model delays and phase among channels. 

\begin{definition}
    The multi-output spectral mixture (MOSM) kernel between channels $i$ and $j$ has the form:
    \begin{equation}
    \begin{aligned}
        k_{ij}(\hat{x})=\sum_{q=i}^{Q}&\alpha_{ij}^{(q)}\exp{(-\frac{1}{2}(\hat{x}+\theta_{ij}^{(q)})^{\top}\Sigma_{ij}^{(q)}(\hat{x}+\theta_{ij}^{(q)}))}\\&\cos{((\hat{x}+\theta_{ij}^{(q)})^{T}\mu_{ij}^{(q)}+\phi_{ij}^{(q)})},
    \end{aligned}
    \end{equation}
    where $\hat{x} = x-x'$ is the difference between input locations, and the hyperparameters $\alpha_{ij}^{(q)},\Sigma_{ij}^{(q)}, \mu_{ij}^{(q)}, \theta_{ij}^{(q)}, \phi_{ij}^{(q)}$ are the magnitude, covariance, mean, time delay and phase delay (respectively) between channels $i$ and $j$. 
\end{definition}

The MOSM kernel exhibits desirable properties for modeling multivariate time series, in particular,  it provides a clear interpretation from a spectral viewpoint through its hyperparameters. However, since this family of kernels stems form Cramér’s theorem, the method is only suited for stationary processes.

\subsection{Harmonizable processes}

To consider a general class of processes beyond stationary ones, we will study the celebrated extension of the stationarity property called harmonizabilty,  originally introduced in \cite{loeve1978probability} for the univariate case, and then in \cite{kakihara1997multidimensional} for the multidimensional case. 

\begin{definition}
    A stochastic process on $\R^{d}$ is weakly harmonizable iff its covariance function can be expressed as:
    \begin{equation}
    \label{harmonizable}
        k(x,x') = \iint_{\R^d\times \R^d}e^{\iu(\omega^{\top}x-\omega'^{\top}x')}F(d\omega,d\omega'),
    \end{equation}
    where $\iu$ is the imaginary unit and $F(d\omega,d\omega')$ is a positive semi-definite bimeasure\footnote{$F$ is a bimeasure iff $F(A,\cdot)$ and $F(\cdot,B)$ are complex measures $\forall A,B \in \mathcal{B}(\R^d)$, but is not necessarily a measure on $\mathcal{B}(\R^d)\otimes\mathcal{B}(\R^d)$ } with finite Fréchet variation, referred to as the spectral bimeasure of the process. A harmonizable process is strongly harmonizable iff its spectral bimeasure is a measure and the integral above coincides with the Lebesgue integral.
\end{definition}

Henceforth, we will refer to strongly harmonizable processes simply as \textit{harmonizable processes}.

\begin{remark}
    When $F$ is absolutely continuous w.r.t. the Lebesgue measure, we denote its Radon-Nikodym derivative as $S = \frac{\partial^2 F}{\partial \omega\partial\omega'}$ and write the covariance as 
    \begin{equation}
        k(x,x')=\iint_{\R^d\times\R^d}e^{\iu(\omega ^{\top}x-\omega'^{\top}x')} S(w,w')d\omega d\omega',
    \end{equation}
    we refer to $S$ as the generalized spectral density of the process in analogy to the spectral density of stationary processes.  
\end{remark}
\begin{remark}
    Notice that $S(\omega,\omega')$ is also a covariance function \cite{hurd1973testing}, and measure the interaction between the $\omega$ and $ \omega'$, thus, we can truly interpret the variables $\omega$ and $\omega'$ as frequencies. This notion of correlation between frequencies is what gives harmonizable processes the property of modeling non-stationary processes.
\end{remark}

\begin{remark}
    Observe that the harmonizable concept is a consistent extension of stationary processes in the sense that when the measure $F$ concentrates on its diagonal $\omega = \omega'$, eq.~\eqref{harmonizable} collapses to eq.~\eqref{Bochner} in Bochner Thm. 
\end{remark}

It is worth noting that the harmonizable processes as presented above define a much larger class than that of stationary processes. In fact, \cite{yaglom1987correlation} noticed that the only processes with continuous bounded kernels that are not harmonizable are, in his own words:
\vspace{-0.5em}

\centerline{\textit{rather complicated and have some unusual,}}
\centerline{\textit{ even pathological, properties.}}
\vspace{0.25em}

More recently \cite{samo2017advances}, who studied the universality of the harmonizable kernels, showed that:
\begin{prop}
     The family of harmonizable kernels defined on $\R^{d}\times\R^{d}$ is pointwise dense in the family of all complex-valued continuous bounded kernels defined on $\R^{d}\times\R^{d}$.
\end{prop}

Moreover, the concept of harmonizability can also be extended to the multivariate case as follows.

\begin{theorem}[\cite{kakihara1997multidimensional}]
    \label{th:mo-harmonizable}
    A family $\{k_{ij}(x,x')\}_{i,j=1}^{m}$ of complex-valued functions on $\R^d$ are the covariance functions of a harmonizable multivariate stochastic process on $\R^d$ if and only if they admit the following representation:
     \begin{equation}
         k_{ij}(x,x')=\iint_{\R^d\times \R^d}e^{\iu(\omega^{\top}x-\omega'^{\top}x')}F_{ij}(d\omega,d\omega'),
    \end{equation}
    where the matrix spectral measure $F=[F_{ij}(A,B)]$ is such that $\forall i,j, F_{ii}$ is positive semi-definite, and symmetric, that is, $F_{ij}(A,B)=\overline{F}_{ji}(B,A)$.
\end{theorem}
\begin{remark}
    In the same manner as in the univariate case, if $F$ is absolutely continuous w.r.t. the Lebesgue measure, we denote its Radon-Nikodym derivative as $S = \frac{\partial^2 F}{\partial \omega\partial\omega'}$, the generalized spectral density of the process and we have 
    \begin{equation}
        k_{ij}(x,x')=\iint_{\R^d\times\R^d}e^{\iu(\omega ^{\top}x-\omega'^{\top}x')} S_{ij}(w,w')d\omega d\omega'.
    \end{equation}
\end{remark}


\section{MULTI-OUTPUT HARMONIZABLE SPECTRAL MIXTURE}
\label{proposal}


Following the idea behind the MOSM kernel,  we propose a family of Hermitian positive-definite complex-valued functions $\{S_{ij}\}_{i,j=1}^{M}$ that satisfy the Theorem \ref{th:mo-harmonizable} to use them as building blocks for a generalized cross-spectral densities.

In order to model the relationship among channels, we support the proposed family of Hermitian positive-definite complex-valued functions on its Cholesky decomposition, recalling that every complex-valued positive-definite $m\times m$ matrix $S$ can be decomposed as $S(\omega,\omega') = R^{H}(\omega,\omega')R(\omega,\omega')$, where $R\in\C^{Q\times m}$, with $Q\in\mathbb{N}$ the rank of the decomposition. For ease of understanding, we choose $Q=1$, the case for arbitrary $Q$ is shown at the end of the section. Since, the $(i,j)$ entry of $S(\omega,\omega')$ can be expressed as $S_{ij}(\omega,\omega')=\overline{R}_i(\omega,\omega')R_j(\omega,\omega') $, $\forall i,j=1,\ldots,m$, thus, we choose:
\begin{equation}
\begin{aligned}
    &R_i(\omega,\omega') = w_{i}\underbrace{\exp{\left(-\dfrac{1}{4l_{i}^{2}}\|\hat{\omega}\|^{2}\right)}}_{(\star)}\times
    \\&\underbrace{\exp{\left(-\dfrac{1}{4}(\overline{\omega}-\mu_{i})^{\top}\Sigma_{i}^{-1} (\overline{\omega}-\mu_{i})-\iu(\theta_{i}^{\top}\overline{\omega}+\phi_{i})\right)}}_{(\bullet)},
\end{aligned}
\end{equation}
where $\hat{\omega} = \omega-\omega',\  \overline{\omega} = (\omega+\omega')/2$, and $w_i,\phi_i \in \R$, $\theta_i,\mu_i\in\R^n$,  $\Sigma_{i} = \diag([\sigma_{i1}^{2},...,\sigma_{in}^{2}])\in\R^{n\times n}$ are hyperparameters.

The intuition behind this choice is that the $(\star)$ term controls the correlation between the frequencies: two frequencies that are further away from one another are less correlated. On the other hand, the $(\bullet)$ component models the importance of each frequency. We parametrize both components as square exponential (SE) functions with complex argument, since: i) they are closed under multiplication and anti-Fourier transform, meaning that the resulting kernel is explicit and ii) they can match continuous power spectra to a desired degree of accuracy.

Therefore, selecting this parametrization for $\{R_{i}\}_{i=1}^{m}$ we have that $\{S_{ij}\}_{i,j=1}^{m}$ are given by:
\begin{equation}
    \begin{aligned}
            S_{ij} = &w_{ij}\exp{\left(-\dfrac{1}{2l_{ij}^{2}}\|\hat{\omega}\|^{2}\right)}\times\\
            &\exp{\left(-\dfrac{1}{2}(\overline{\omega}-\mu_{ij})^{\top}\Sigma_{ij}^{-1} (\overline{\omega}-\mu_{ij})\right)}\times\\
            &\exp(-\iu(\theta_{ij}^{\top}\overline{\omega}+\phi_{ij})).
    \end{aligned}
\end{equation}

Observe that this is a decaying square exponential---due to the factor $\exp\left(-\frac{1}{2l_{ij}^{2}}\|\hat{\omega}\|^{2}\right)$---and the channel parameters obey the following relationships:\\\\
$\bullet$ covariance: $\Sigma_{ij} = 2\Sigma_{i}(\Sigma_{i}+\Sigma_{j})^{-1}\Sigma_{j}$\\
$\bullet$ mean: $\mu_{ij} = (\Sigma_{i}+\Sigma_{j})^{-1} (\Sigma_{i}\mu_{j}+\Sigma_{j}\mu_{i})$\\
\mbox{$\bullet$ magnitude: $w_{ij} = w_{i}w_{j}\exp{\left(\frac{(\mu_{i}-\mu_{j})^\top(\Sigma_{i}+\Sigma_{j})^{-1}(\mu_{i}-\mu_{j})}{4}\right)}$}\\
$\bullet$ delay: $\theta_{ij}=\theta_i - \theta_j$\\
$\bullet$ phase: $\phi_{ij} = \phi_i-\phi_j$\\
$\bullet$ length-scale: $l_{ij}^{2} = 2l_{i}^{2}l_{j}^{2}(l_{i}^{2}+l_{j}^{2})^{-1}$.\\

Observe that $S_{ij}$ is a \textit{locally-stationary kernel}, a concept coined by \cite{silverman1957locally}, which refers to kernels that can be expressed as a product of a stationary kernel and a non-negative function. Therefore we can assure that each $S_{ij}$ is a positive-definite complex-valued function. Additionally, since $S$ was designed through the product $S=R^{H}R$, it is by construction a positive-definite matrix thus fulfilling Theorem \ref{th:mo-harmonizable}. Finally, as we are interested in real-valued GPs, for which the covariance kernel is also required to be real-valued,  we make $S_{ij}$ symmetric by reassigning:
\begin{equation}
    S_{ij}(\omega,\omega')\longleftarrow \frac{1}{2}(S_{ij}(\omega,\omega')+S_{ij}(-\omega,-\omega')),
\end{equation}
this guarantees symmetry and real values in the diagonal as the complex terms cancel each other. Therefore, the kernel obtained by taking the integral of the symmetrised spectral density is:
\begin{equation}
    \label{MOHSM1}
    \begin{aligned}
   k_{ij}(x,x') = &\alpha_{ij}\exp{(-\frac{1}{2}(\hat{x}+\theta_{ij})^{\top}\Sigma_{ij}(\hat{x}+\theta_{ij}))}\times \\
   &\cos{((\hat{x}+\theta_{ij})^{T}\mu_{ij}+\phi_{ij})}\exp{(-\frac{1}{2l_{ij}^{2}}\|\overline{x}\|^{2})},
    \end{aligned}
\end{equation}
where $\overline{x} = \frac{x+x'}{2}$, $\hat{x} = x-x'$ and the magnitude $\alpha_{ij}=w_{ij}(2\pi)^{n}|\Sigma_{ij}|^{1/2}l_{ij}$.


From the kernel and spectral expressions we can interpret the parameters of the constructed kernel as follows:
\begin{itemize}
    \item The spectral mean $\mu_i$ represents the main frequency
    \item The spectral covariance $\Sigma_i$ represents the uncertainty of the distribution in the spectrum
    \item The cross spectral delay $\theta_{ij}$ serves as the time delay between channels
    \item The cross spectral phase $\phi_{ij}$ provides the difference in phase between channels
    \item The spectral length-scale $l_i$  controls the correlation between the frequencies.
\end{itemize}

One drawback of the presented formulation is that, due the last exponential term in eq.~\eqref{MOHSM1}, the proposed kernel vanishes outside the origin with length-scale $l_i$. We address this limitation by placing input shifts, which will allow us to control in which parts of the input domain the kernel is activated. We can include the input shifts in our formulation by multiplying our initial matrix $S$ by $\exp(-x_p(w-w'))$, which comply with the properties needed for Theorem \ref{th:mo-harmonizable} to hold. Putting together all the aforementioned components, the kernel is defined as follows:
\begin{equation*}
    \begin{aligned}
        k_{ij}(x,x') = &\alpha_{ij}\exp{(-\frac{1}{2}(\hat{x}+\theta_{ij})^{\top}\Sigma_{ij}(\hat{x}+\theta_{ij}))}\times\\
        &\cos{((\hat{x}+\theta_{ij})^{T}\mu_{ij}+\phi_{ij})}\times\\
        &\exp{(-\frac{1}{2l_{ij}^{2}}\|\overline{x}-x_p\|^{2})}.
    \end{aligned}
\end{equation*}

Lastly, for the general case we can expand the kernel to an arbitrary rank matrix $Q$ by taking $S$ as a sum of $Q$ of these matrices of rank 1, and considering $P$ input shifts. This yields the final expression for the proposed kernel, termed MOHSM:
\begin{definition}
    \label{def:MOHSM}
    The Multi-Output Harmonizable Spectral Mixture (MOHSM) kernel has the form:
    \begin{equation}
    \begin{aligned}
        k_{ij}(\hat{x},\overline{x}) = \sum_{p=1}^{P}\sum_{q=1}^{Q_p}&\alpha_{ij}^{(q)}\exp(-\frac{1}{2}(\hat{x}+\theta_{ij}^{(q)})^{\top}\Sigma_{ij}^{(q)}(\hat{x}+\theta_{ij}^{(q)}))\times\\
        &\cos{((\hat{x}+\theta_{ij}^{(q)})^{\top}\mu_{ij}^{(q)}+\phi_{ij}^{(q)})}\times\\
        &\exp(-\frac{l_{ij}^{2(p)}}{2}\|\overline{x}-x_p\|^{2}),
    \end{aligned}
    \end{equation}
    where recall that $\hat{x} = x-x'$ and $\overline{x}=\frac{x+x'}{2}$, $P$ is the number of input shifts, $Q_p$ is the number of spectral components for the $p^{th}$ input shift with location $x_p$, $\alpha_{ij}^{(q)}=w_{ij}^{(q)}(2\pi)^{n}|\Sigma_{ij}^{(q)}|^{1/2}l_{ij}^{(q)}$, the super index $(\cdot)^{(q)}$ denotes the parameter of the $q^{th}$ component of the spectral mixture and the super index $(\cdot)^{(p)}$ denotes the parameters for the $p^{th}$ input shift.
\end{definition}

Using $Q_p > 1$ can also be justified by the latent-factor construction of MOGPs. There, $Q_p$ denotes the number of latent GPs, and since each of these latent signals has different kernels, the "capacity" of the model is not dominated by the number of channels $m$ but by $Q_p$, which is (proportional) to the amount of the kernel's hyperparameters.


\section{RELATIONSHIP TO PREVIOUS WORK}
\label{sec:previous_work}

Even though the idea of considering a non-stationary kernel derived from the harmonizable processes is not new, all previous attempts are restricted to the single output case. For example, \cite{samo2015generalized} proposed a family of spectral kernels that they prove can approximate any continuous bounded nonstationary kernel which they called Generalized Spectral Kernels. In the same line, \cite{shen2019harmonizable} proposed the harmonizable spectral mixture (HSM) kernel which is also a family derived from mixture models on the generalized spectral representation. \cite{remes2017nonstationary} presented a non-stationary kernel based on the idea of harmonizable processes and parametrized the frequencies, length scales, and mixture weights as Gaussian processes. Our work can be seen as a multivariate extension of the family proposed by \cite{shen2019harmonizable} with expressive cross-covariance functions.

In general, classical MOGP approaches (such as the Linear Model of Corregionalization and the Intrinsic Corregionalization Model) can represent non-stationary processes, since they model the cross-correlation functions as a linear combination of the auto-correlation functions, thus choosing non-stationary auto-correlations leads to a non-stationary multivariate process. The issue with these formulations is they force the auto-covariance and the cross-covariance to have similar behavior. Also, these methods based on linear mixtures are not able to introduce temporal correlation other than those of the latent GP component. Furthermore, by modeling the cross-correlation in this fashion, the interpretability of the learned dependence is almost null. Though MOSM solves these previous problems, adding interpretability of the dependencies and not imposing similar behaviors through different channels, is restricted to stationary cases by construction. 

\begin{figure}[t]
\centering
\begin{tikzpicture}
        \draw[thick,<->] (-2.5,0) -- (2.5,0);
        \draw[thick,<->] (0,-2.5) -- (0,2.5);
        \node at (2.5,0)[below right,rotate=45]  {Non Stationary};
        \node at (-2.5,0)[above=2pt,rotate=45]  {Stationary};
        \node at (0,2.5)[above=2pt] {Single Output};
        \node at (0,-2.5)[below=2pt] {Multi Output};
        \node[circle,draw,minimum size=1.5cm] at (1, -1){{\color{red}MOHSM}};
        \node[circle,draw,minimum size=1.5cm] at (-1, -1){MOSM};
        \node[circle,draw,minimum size=1.5cm] at (1, 1){HSM};
        \node[circle,draw,minimum size=1.5cm] at (-1, 1){SM};
        \end{tikzpicture}
\caption{Relationship between the different spectral kernels} \label{fig:M1}
\end{figure}
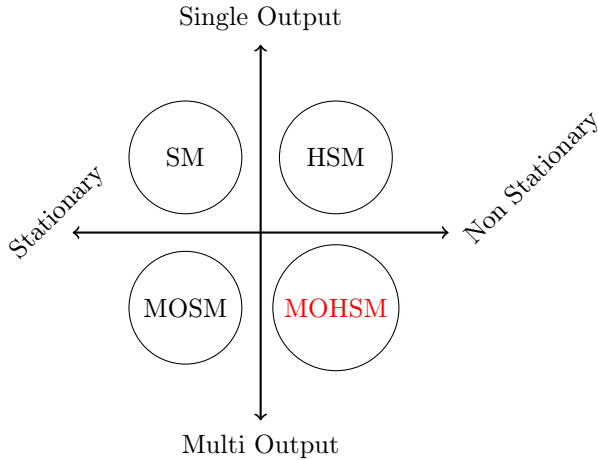

In this way, our work is a generalization of MOSM since the proposed MOHSM can model the correlation between the channels like the MOSM do, but allowing changes in the regimes across time, leading to a more general model.  A natural question that arises in our context is whether the MOHSM can recover its stationary counterpart MOSM. We can notice that when $x_p=0$ and $l_{ij}\to0$, $\forall i,j$ we recover the MOSM kernel, successfully extending the stationary model (more details in the supplementary material). In the frequency domain, this can be seen as the absence of correlation between frequencies, which is the assumption of stationarity. Figure \ref{fig:M1} shows a summary of the relationships of the MOHSM with the different spectral kernels.

It is worth discussing what kind of non-stationarity the proposed MOHSM model can capture. Observe, from Def.~\ref{def:MOHSM}, that each mixture of the proposed kernel is a locally stationary kernel, which means that the proposed kernel is a combination of a component of global structure multiplied by components that describe the local structure of the data. The chosen global structure is an exponential window operating over the regions described by the local structure component. Thus, the kernel can be seen as a union of different regimes that can be disjoint or overlapped, depending on the locations of the input shifts $x_p$ and their lengthscales $l^{(p)}$.

\section{PRACTICAL CONSIDERATIONS}
\label{sec:practical_cons}
\subsection{Training}

Training an MOGP with the MOHSM kernel follows the same procedure as the single output GP, by closed-form maximum likelihood. Since the model is jointly Gaussian, we can concatenate the locations $x\in\R^n$, the channel identifier $i\in\{1,...,M\}$ and the observed value $y\in\R$ into three vectors and express the negative log-likelihood (NLL) as considering the multi-output covariance matrix of all the observed samples. 

The NLL is minimised with respect to all the hyperparameters of the MOHSM including the noise hyperparameter, that is
\begin{equation*}
    \Theta = \{w_i^{(q)},\mu_i^{(q)},\Sigma_i^{(q)},\theta_i^{(q)},\phi_i^{(q)},\sigma_{i,\text{n}}^{(q)},l_{i}^{(p)},x_p\}_{i=1,q=1,p=1}^{M,Q_p,P}.    
\end{equation*}
Once the optimization is concluded, computing the predictive posterior in the proposed model follows the standard GP procedure as well.

In order to make the optimization more stable we reparametrized the MOHSM kernel using $\gamma_{ij} = l_{ij}^{-1}$. This will allow us to express the global component of the MOHSM in a more amenable form to find models where $\gamma_{ij}$ approaches zero rather than $l_{ij}$ going to infinity.

\subsection{Parameter initialization}
The MOHSM kernel, akin to the MOSM and other spectral mixture kernels, is challenging to train as its likelihood is particularly susceptible to local maxima, and sensible to initial conditions: a poor choice of initial condition will lead to suboptimal solutions. In this context, we suggest a way of initializing the hyperparameters based on the interpretation of each parameter from a spectral viewpoint. 

First, we set the centers and the length-scale of each component. We suggest, after choosing how many different centers consider, to place them equidistant and cover the input space with windows of the same length-scale. Now, to initialize the parameters of each regime we recommend a scheme based on the estimated power spectrum of the (windowed) data. Recently,  \cite{cuevas2020multi}  proposed an initialization scheme for the SM kernel and the MOSM kernel, which leads to consistent and better results. This concept estimates the power spectral density (PSD) of the available data to then use it to obtain initial values of the parameters, which have a spectral interpretation. Since the MOHSM is a spectral-inspired  kernel, we can rely on the above idea to initialize its hyperparameters too. The MOGPTK toolkit \cite{de2020gaussian} has implemented this initialization strategy for spectral mixture kernels, in particular, using the Bayesian non parametric spectral estimation method by \cite{tobar2018bayesian}.  In our experience, the initial conditions of the time delay and phase parameters are best set to zero, thus making a initial assumption that there is no input-delay or phase-delay between channels, leaving to the optimization process to find the non-zero delay and phase if the data reveals so.


\section{EXPERIMENTS}
\label{experiments}
We tested the proposed MOHSM kernel in different settings. First we learned a synthetic three-output GP, then we applied MOHSM to two real-world scenarios: a dataset comprising series of gold and oil prices, the NASDAQ and the USD index (henceforth referred to as GONU), and electroencephalography (EEG) data which are known to have dependencies among frequencies. 

Since the MOSM kernel has shown better results than the others stationary MOGP kernels, such as the CONV and the CSM, and it is conceptually more general than them, we only considered MOSM as our stationary benchmark. Additionally, we compared MOHSM against 2 other non-stationary MOGP kernels: an independent non-stationary kernel per channel, and non-stationary linear model of coregionalization. The selected non-stationary kernel for both non-stationary MOGP kernels is defined as follow:
\begin{equation}
\begin{aligned}
    k(x,x')=\sum_{q=1}^{Q}&w_{q}\exp\left(-\frac{1}{2l_{q}^{2}}\left\|\dfrac{x+x'}{2}-c_q\right\|^2\right)\\
    &\exp\left(-\frac{1}{2}\tau^{\top}\Sigma_{q}\tau\right)\cos(\mu_{q}^{\top}\tau),
\end{aligned}
\end{equation}

where $\tau = x-x'$.

The above defined kernel is from the family of harmonizable mixture kernels proposed by Shen et al. \cite{shen2019harmonizable}, which can be seen as a SM kernel where each component is windowed and centered in $c_q$. Therefore, we refer to the  independent harmonizable HM (HSM) kernel per channel simply as HSM, and the non-stationary linear model of coregionalization using the HSM kernel is called HSM-LMC.

All models were implemented for our experiments based on the architecture of MOGPTK for GPU-accelerated ML-training of GPs \cite{mogptk}. Moreover, the MOHSM kernel is now part of MOGPTK, and the code for the experiments can be found in its Github repository.\footnote{\href{https://github.com/GAMES-UChile/mogptk}{https://github.com/GAMES-UChile/mogptk}} The simulations were executed on a Intel Core i7 - 7500U 2.7 GHz CPU with 8 GB of RAM and a 940MX GPU.

\subsection{Learning derivatives and delayed signals}

This experiment demonstrates the expressiveness of the MOHSM by using it to recover the auto and cross covariance of a nonstationary MOGP. We simulated the following three signals: a sample $f$ from a GP with a non stationary kernel and zero mean, its derivative and a delayed version of the GP. The relevance of this experiment stems from the fact that the true covariance and cross covariance of the mentioned process are known explicitly, thus we can test the performance of the model. 

We produced $N=500$ samples in the interval $[-20,20]$ for each channel. For the experiment, the derivative was computed numerically and we removed observations between $[-10,-5]$ for the derivative and between $[-5,5]$ for the delayed signal. We randomly split the dataset into 70\% for training and 30\% for testing. We used a HSM kernel with 2 mixtures, one centered in $-20$, and the other centered in $20$. In order to measure the performance of the models, we will use the distance between the correlation matrix, which is a metric defined by \cite{herdin2005correlation} that measures the similarity between two covariance matrices, and is defined as follows:

\begin{definition}
The correlation matrix distance (CMD) \cite{herdin2005correlation} is the distance between two correlation matrices $K_1$ and $K_2$ as defined by
\begin{equation}
    CMD(K_1,K_2) = 1- \dfrac{\Tr(K_1\cdot K_2)}{\|K_1\|\cdot \|K_2\|},
\end{equation}
where the norm is the Frobenius norm and $\Tr$ denotes the trace.
\end{definition}

\begin{figure}[h]
\centering
\includegraphics[scale=0.24]{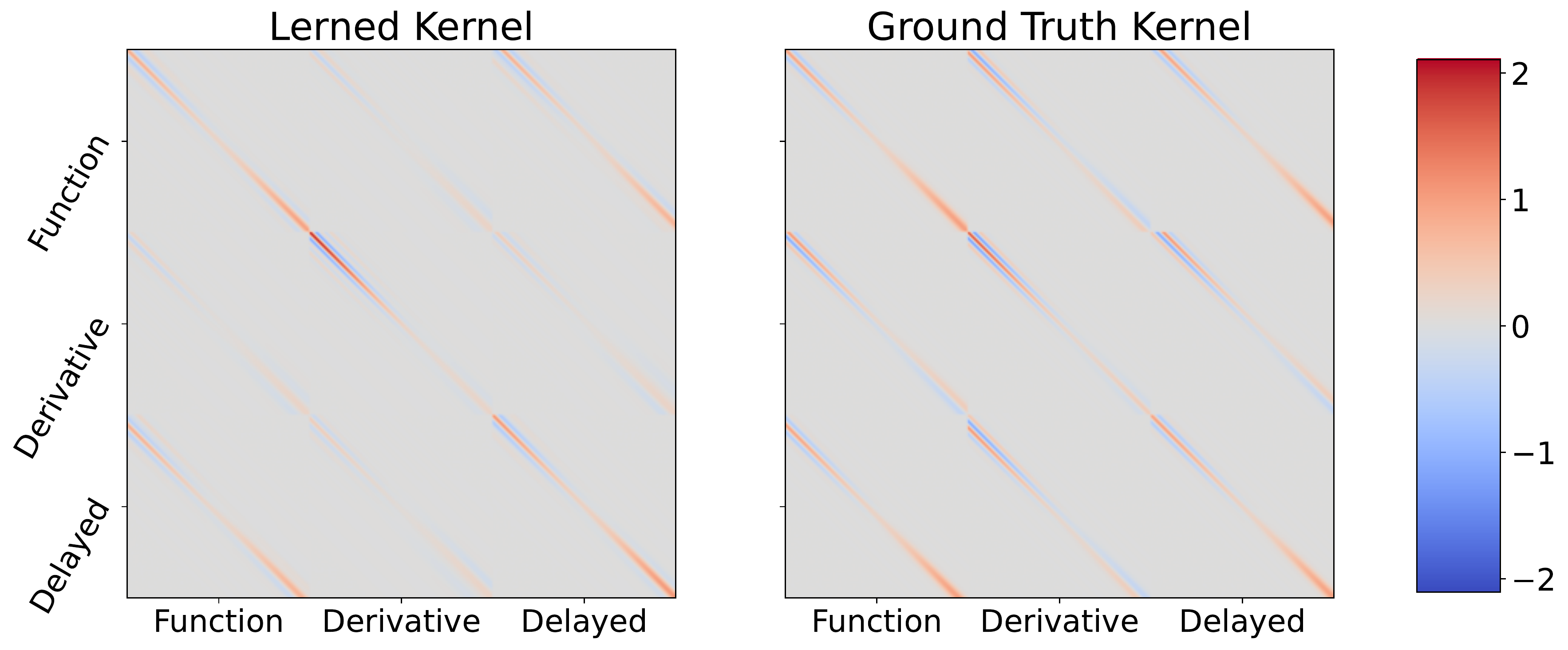}
\captionof{figure}{\textbf{Left:} learned covariance function by MOHSM kernel. \textbf{Right:} ground truth covariance function of the synthetic dataset.}
\label{fig:cov}
\end{figure}
\begin{table}[h]
    \centering
  \caption{Performance indices for the synthetic dataset using the correlation matrix distance (CMD) over 5 realizations}
  \begin{tabular}{@{}cc@{}}
    \toprule
    Method & CMD            \\ 
    \midrule
    MOSM       & 0.85 ± 0.01\\
    HSM        & 0.86 ± 0.00\\
    HSM-LMC    & 0.80 ± 0.00\\
    MOHSM     & \textbf{0.48 ± 0.12} \\ 
    \bottomrule
    \end{tabular}
    \label{tab:synthetic}
\end{table}
 \begin{figure}[h!]
    \centering
    \includegraphics[scale=0.3]{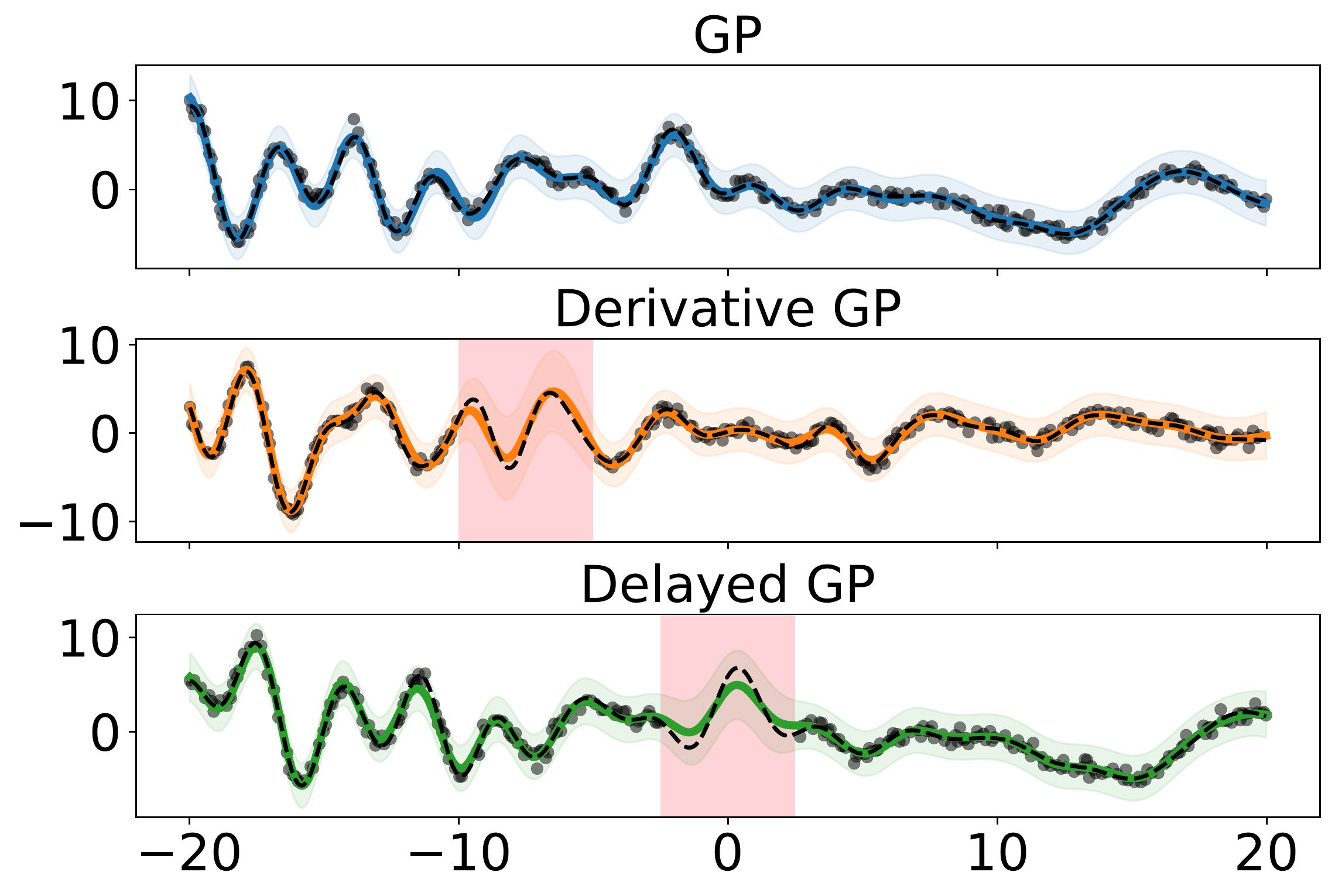}
    \caption{Synthetic data set with the trained \textbf{MOHSM} kernel. Training points are shown in black, dashed lines are the ground truth and the colour coded lines are the posterior means. The coloured bands show the 95\% confidence intervals. The red shaded areas mark the data imputation ranges.}
    \label{fig:synthetic}
\end{figure}

\begin{figure*}[t]
    \centering
    \includegraphics[scale=0.3]{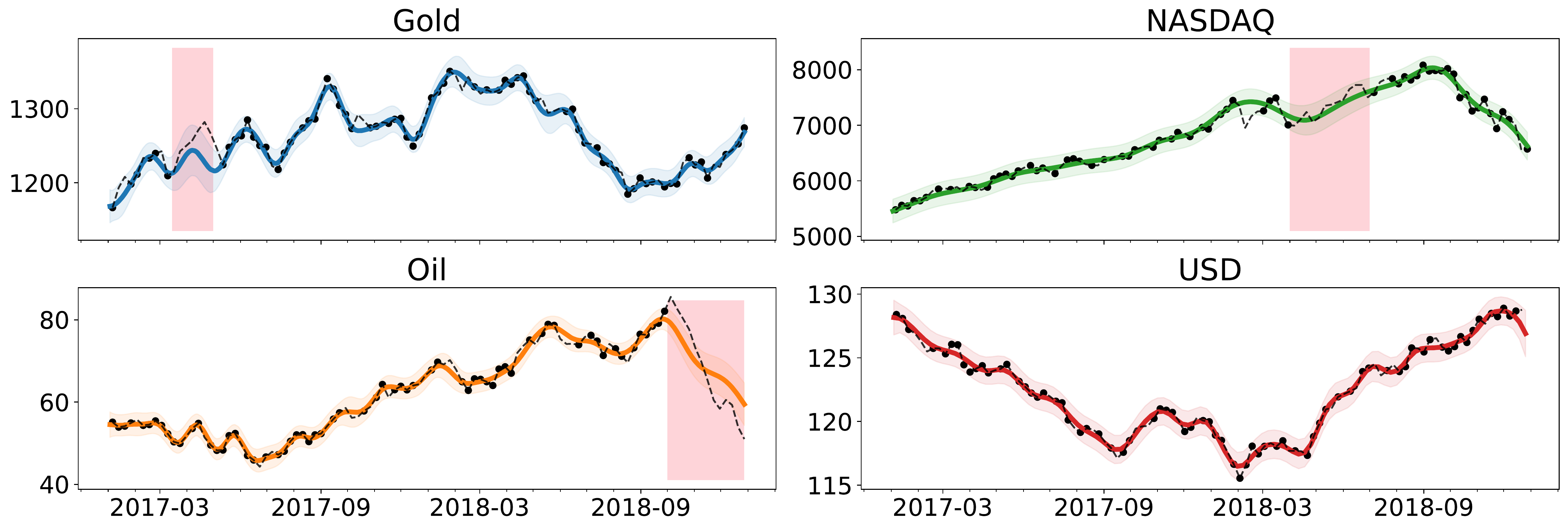}
    \caption{GONU data set with the trained MOHSM kernel. Training points are shown in black, dashed lines are the ground truth and the colour coded lines are the posterior means. The coloured bands show the 95\% confidence intervals. The red shaded areas mark the data imputation ranges}
    \label{fig:GONU}
\end{figure*}

Figure \ref{fig:cov} shows the learned covariance function by MOHSM kernel compared to the ground truth covariance for the synthetic dataset. In this figure we observe that the MOHSM kernel is able to recover almost perfectly the ground truth kernel, learning both regimes and the delay of the third channel. Figure \ref{fig:synthetic} shows that the MOHSM model was able to accurately recover the auto and cross covariance of the reference GP and the delayed version, yet it exhibits slightly larger error bars for the derivative. Among the 4 models considered, Table \ref{tab:synthetic} confirms that the MOHSM largely outperforms the benchmarks in terms of the correlation matrix distance. Notably, in this example MOHSM achieved satisfactory results learning the cross-covariance between the GP and the delayed version, and autocovariances without prior information about the delays.

\subsection{GONU dataset}

In previous work, \cite{de2020gaussian} proved the capability of MOSM to impute and predict financial observation by learning the relationships among financial time series. Although MOSM has performed successfully in these applications, the underlying assumption of stationarity is too strong, since financial series are known to be nonstationary. For example, \cite{joyeux1980harmonizable} have found that, with new housing starts, the high and low frequency components were inter-correlated, and thus the series might be mismodeled by assuming stationarity.

In order to validate hypothesis on nonstationarity, we applied MOHSM to one of the \cite{de2020gaussian} experiments, a dataset comprising series of gold and oil prices, the NASDAQ and the USD index, between January 2017 and December 2018 with a weekly frequency, so as to account for the known stylized facts of financial series\footnote{The dataset can be found in the websites: \href{https://www.eia.gov/dnav/pet/hist/RBRTEd.htm}{www.eia.gov/dnav/pet/hist/RBRTEd.htm}, \href{https://finance.yahoo.com}{finance.yahoo.com} \& 
\href{https://fred.stlouisfed.org/series/TWEXB.
}{fred.stlouisfed.org/series/TWEXB}}. To simulate missing data we removed regions in each channel. We trained a MOHSM kernel with 4 mixtures on 385 points, and tested on 446 points.

Figure \ref{fig:GONU} shows that the MOHSM gives an accurate fit to the dataset, where practically all the data are within the predicted confidence interval. Moreover, in the regions where the data were removed, the model was able to predict within an acceptable precision. The performance of the other methods can be found in the supplementary material.Table \ref{tab:GONU} shows a quantitative comparison  of different models against the MOHSM. We performed 5 trials per trained model and reported the mean and the standard deviation of the mean absolute percentage error (MAPE) and root mean square error (RMSE) 

\begin{table}[t]
\caption{Performance indices for the GONU dataset using the mean absolute percentage error (MAPE), root mean square error (RMSE) and negative log likelihood (NLL) over 5 realizations}
\scriptsize
\centering
\begin{tabular}{@{}cccc@{}}
\toprule
Method & MAPE        & RMSE        & NLL \\ 
\midrule
MOSM    & 3.05 ± 0.27   & 49.55 ± 5.79  & -155.147 ± 9.09\\
HSM     & 3.20 ± 0.37   & 43.54 ± 2.67  & -99.97 ± 3.58 \\
HSM-LMC & 2.49 ± 0.40   & 64.20 ± 12.11 & -111.43 ± 20.84 \\
MOHSM   & \textbf{1.67 ± 0.16}  & \textbf{40.44 ± 4.74} & \textbf{-164.83 ± 10.15}\\ 
\bottomrule
\end{tabular}
\label{tab:GONU}
\end{table}

\setlength{\tabcolsep}{2pt}
\renewcommand{\arraystretch}{1}

\begin{table*}[t]
\centering
\scriptsize
\begin{tabular}{@{}cccccccccc@{}}
\toprule
Model & Fp1        & Fp2         & Fz         & Cz         & T3         & T4        & O1    & O2  & \textbf{Overall}\\ 
\midrule
MOSM  & 0.16 ± 0.01 & 0.12 ± 0.01         & 0.16 ± 0.01 & \textbf{0.12 ± 0.00} & \textbf{0.13 ± 0.02} & 0.16 ± 0.02          & \textbf{0.17 ± 0.00} & 0.12 ± 0.02 & 0.14 ± 0.01\\
HSM   & 0.16 ± 0.01 & 0.17 ± 0.01 & 0.17 ± 0.00 & 0.12 ± 0.01 & 0.38 ± 0.01 & 0.24 ± 0.01 & 0.27 ± 0.00 & 0.29 ± 0.01 & 0.22 ± 0.01\\
HSM-LMC   & 0.16 ± 0.01 & \textbf{0.11 ± 0.01} & 0.14 ± 0.01 & 0.13 ± 0.01 & 0.16 ± 0.02 & 0.15 ± 0.01 & 0.33 ± 0.06 & 0.17 ± 0.03 & 0.17± 0.02\\
MOHSM & \textbf{0.14 ± 0.01}          & 0.13 ± 0.02 & \textbf{0.13 ± 0.01}          & \textbf{0.12 ± 0.00}           & 0.14 ± 0.01      & \textbf{0.13 ± 0.00} & 0.18 ± 0.02  & \textbf{0.09 ± 0.01}   &    \textbf{0.13 ± 0.01}    \\ 
\bottomrule
\end{tabular}
\caption{Performance for the EEG dataset of each channel using the normalized mean absolute error (nMAE) over 5 realisations}
\label{tab:EEG}
\end{table*}


\subsection{EEG dataset}

In the field of neuroscience, practitioners need to accurately model encephalography (EEG) data so as to correctly study brain states, in particular, the frequency-based perspective is the standard in multivariate EEG analysis \cite{gorrostieta2019time}. Since the MOHSM is able to learn the interaction between frequencies, the proposed model is well suited to these challenge.

We tested MOHSM on an EEG dataset with 8 channels. We selected a 60-second window resampled at 2 [Hz] and randomly split the data set into 70\% for training and 30\% for testing. We trained on 735 points a MOHSM kernel with 4 mixture, and tested on 315 points.

Table \ref{tab:EEG} shows the results for the  different models considered the EEG experiment, where we performed 5 trials per trained model and reported the mean and the standard deviation of the normalized mean absolute error (nMAE), for each channel. 

\section{CONCLUSION}
\label{conclusion}
We have presented the multi-output harmonizable spectral mixture (MOHSM) kernel as a generalization of the well-known multi-output spectral mixture (MOSM) kernel to the non-stationary case. The proposed family of kernels relies upon the concept of harmonizable processes, a rather general class of processes in the sense that it contains stationary processes and a large portion of the non-stationary processes. The resulting kernel, termed MOHSM,  provides flexibility to model both stationary and non-stationary processes while maintaining the desired properties of the MOSM: a clear interpretation of the parameters from a spectral viewpoint, and flexibility in each channel. Furthermore, we have implemented a parameter initialization scheme to overcome the sensibility of the MOHSM to initial conditions. We have showed that our method can effectively model non-stationary data and is a sound extension of the MOSM kernel. Future work includes considering more complex spectral densities instead of Gaussian functions, this would allow us to prescind of the infinite differentiability requirement of sampled functions assumed by spectral mixture kernels. A sparse implementation of MOSHM is also part of the future work, yet that is a challenge in its own since the relationship between the locations of inducing inputs in the multichannel case has not been thoroughly studied so far. Finally, we hope our work catalyzes interest in the harmonizable processes and their  role on multichannel models.

\subsubsection*{Acknowledgments}
We thank Taco de Wolff for his advice using the MOGPTK toolbox and Jou-Hui Ho for insightful discussions about the EEG experiment. This work was funded by Google, Fondecyt-Regular 1210606, ANID-FB210005 (CMM) and ANID-FB0008 (AC3E).

\bibliographystyle{apalike}
\bibliography{bibliography.bib}

\clearpage
\appendix

\thispagestyle{empty}

\onecolumn \makesupplementtitle

\section{DERIVATION OF THE MOHSM KERNEL}
    Consider the following cross spectral density:
    \begin{equation}
    S_{ij} = w_{ij}e^{\left(-\frac{1}{2l_{ij}^{2}}\|\hat{\omega}\|^{2}\right)}e^{\left(-\frac{1}{2}(\overline{\omega}-\mu_{ij})^{\top}\Sigma_{ij}^{-1} (\overline{\omega}-\mu_{ij})\right)}e^{(-\iu(\theta_{ij}^{\top}\overline{\omega}+\phi_{ij}))}\underbrace{e^{(-\iu \hat{\omega}^{\top}x_p})}_{\text{(input shift)}},
    \end{equation}
    where $\hat{\omega} = \omega-\omega', \overline{\omega} = \frac{\omega+\omega'}{2}$, $w_{ij},\phi_{ij} \in \R$, $\theta_{ij},\mu_{ij},x_p\in\R^n$ and $\Sigma_{ij} = \diag([\sigma_{ij1}^{2},...,\sigma_{ijn}^{2}])\in\R^{n\times n}$. We calculate the inverse generalized Fourier transform of the spectral densities $S_{ij}(\omega,\omega')$ above to obtain the multivariate covariance function:
    \begin{align*}
        k_{ij}(x,x') &= \iint_{\R^n\times\R^n}e^{\iu (\omega^{\top}x-\omega^{\prime T})}S_{ij}(\omega,\omega')d\omega d\omega'\\
        &= \iint_{\R^n\times\R^n}e^{\iu \left(\left(\frac{\omega+\omega'}{2}\right)^{\top}(x-x')+(\omega-\omega')^{\top}\left(\frac{x+x'}{2}\right)\right)}S_{ij}(\omega,\omega')d\omega d\omega'\\
        &= \iint_{\R^n\times\R^n}e^{\iu \left(\overline{\omega}^{\top}\tau +\hat{\omega}^{\top}\overline{x}\right)}S_{ij}(\omega,\omega')d\omega d\omega'  \quad \left(\text{defining } \tau = x-x'\text{ and } \overline{x}=\frac{x+x'}{2}\right)\\
        &=\iint_{\R^n\times\R^n} e^{\iu \left(\overline{\omega}^{\top}\tau +\hat{\omega}^{\top}\overline{x}\right)}w_{ij} e^{\left(-\frac{1}{2l_{ij}^{2}}\|\hat{\omega} \|^{2}\right)}e^{\left(-\frac{1}{2}(\overline{\omega}-\mu_{ij})^{\top}\Sigma_{ij}^{-1}(\overline{\omega}-\mu_{ij})\right)}e ^{(-\iu(\theta_{ij}^{\top}\overline{\omega}+\phi_{ij}))}e^{(-\iu \hat{\omega}^{\top}x_p})d\omega d\omega'\\
         &=w_{ij}\iint_{\R^n\times\R^n} e^{\left(-\frac{1}{2l_{ij}^{2}}\|\hat{\omega}\|^{2}+\iu\hat{\omega}^{\top}(\overline{x}-x_p)\right)}e^{\left(-\frac{1}{2}(\overline{\omega}-\mu_{ij})^{\top}\Sigma_{ij}^{-1}(\overline{\omega}-\mu_{ij})-\iu(\theta_{ij}^{\top}\overline{\omega}+\phi_{ij})+\overline{\omega}^{\top}\tau\right)}d\omega d\omega'\\
         &=w_{ij}\iint_{\R^n\times\R^n} e^{\left(-\frac{1}{2l_{ij}^{2}}\|\hat{\omega}\|^{2}+\iu\hat{\omega}^{\top}(\overline{x}-x_p)\right)}e^{\left(-\frac{1}{2}(\overline{\omega}-\mu_{ij})^{\top}\Sigma_{ij}^{-1}(\overline{\omega}-\mu_{ij})-\iu(\theta_{ij}^{\top}\overline{\omega}+\phi_{ij})+\overline{\omega}^{\top}\tau\right)}d\overline{\omega} d\hat{\omega}\\
         &=w_{ij}\int_{\R^n}e^{\left(-\frac{1}{2l_{ij}^{2}}\|\hat{\omega}\|^{2}+\iu\hat{\omega}^{\top}(\overline{x}-x_p)\right)}d\hat{\omega}
        \int_{\R^n}e^{\left(-\frac{1}{2}(\overline{\omega}-\mu_{ij})^{\top}\Sigma_{ij}^{-1}(\overline{\omega}-\mu_{ij})-\iu(\theta_{ij}^{\top}\overline{\omega}+\phi_{ij})+\overline{\omega}^{\top}\tau\right)}d\overline{\omega}\\
        &=w_{ij}\int_{\R^n}e^{\left(-\frac{1}{2l_{ij}^{2}}\|\hat{\omega}\|^{2}+\iu\hat{\omega}^{\top}(\overline{x}-x_p)\right)}d\hat{\omega}
        \int_{\R^n}e^{\left(-\frac{1}{2}\overline{\omega}^{\top}\Sigma_{ij}^{-1}\overline{\omega}-(\Sigma_{ij}^{-1}\mu_{ij}+\iu(\tau+\theta_{ij}))^{\top}\overline{w}-\frac{1}{2}\mu_{ij}^{\top}\Sigma_{ij}^{-1}\mu_{ij}+\iu\phi_{ij}\right)}d\overline{\omega}\\
         &=\alpha_{ij}e^{-\frac{l_{ij}^{2}}{2}\|\overline{x}-x_p\|^2}e^{\left(\frac{1}{2}(\Sigma_{ij}^{-1}\mu_{ij}+\iu(\tau+\theta_{ij}))^{\top}\Sigma_{ij}(\Sigma_{ij}^{-1}\mu_{ij}+\iu(\tau+\theta_{ij}))-\frac{1}{2}\mu_{ij}^{\top}\Sigma_{ij}^{-1}\mu_{ij}+\iu\phi_{ij}\right)}\\
         &=\alpha_{ij}e^{-\frac{l_{ij}^{2}}{2}\|\overline{x}-x_p\|^2}e^{\left(-\frac{1}{2}(\tau+\theta_{ij})^{\top}\Sigma_{ij}(\tau+\theta_{ij})\right)}e^{\left(\iu(\tau+\theta_{ij})^{\top}\mu_{ij}+\phi_{ij}\right)}.
    \end{align*}
    Now, in order to obtain a real-valued covariance function $k_{ij}(x,x')$ we take the real part of the above covariance, which is equivalent to \textit{symmetrize} the spectral densities $S_{ij}(\omega,\omega')$. Thus, we set:
    \begin{equation*}
        k_{ij}(x,x') = \alpha_{ij}e^{-\frac{l_{ij}^{2}}{2}\|\overline{x}-x_p\|^2}e^{\left(-\frac{1}{2}(\tau+\theta_{ij})^{\top}\Sigma_{ij}(\tau+\theta_{ij})\right)}\cos{\left(\iu(\tau+\theta_{ij})^{\top}\mu_{ij}+\phi_{ij}\right)}.
    \end{equation*}

\section{RECOVERING THE MOSM KERNEL FROM THE MOHSM KERNEL}

Since our goal is to extend the multi-output spectral mixture (MOSM) to the non-stationary case, we study in more detail how to recover the MOSM from the proposed multi-output harmonizable spectral mixture (MOHSM). We notice that considering $x_p=0$ and taking $l_{ij}\to 0$ in MOHSM we recover the MOSM kernel, successfully extending the stationary model. Indeed, in the frequency domain we notice that when $\omega\neq\omega'$:

\begin{equation}
\label{eq1}
    \exp{\left(-\frac{1}{2l_{ij}^{2}}\|\omega-\omega'\|^{2}\right)}\overset{l_{ij}\rightarrow 0}{\longrightarrow} 0.
\end{equation}
On the other hand in the case $\omega=\omega'$ we observe that:
\begin{equation}
\label{eq2}
    \exp{\left(-\frac{1}{2l_{ij}^{2}}\|\omega-\omega'\|^{2}\right)}\overset{l_{ij}\rightarrow 0}{\longrightarrow} 1,
\end{equation}
Combining equations (\ref{eq1}) and (\ref{eq2}) we obtain:
\begin{equation}
    S_{ij}(\omega.\omega')\overset{l_{ij}\rightarrow 0}{\longrightarrow}\delta(\omega-\omega')\exp{\left(-\dfrac{1}{2}(\overline{\omega}-\mu_{ij})^{\top}\Sigma_{ij}^{-1} (\overline{\omega}-\mu_{ij})\right)}\exp(-\iu(\theta_{ij}^{\top}\overline{\omega}+\phi_{ij})) = \hat{S}(\omega,\omega'),
\end{equation}

where $\delta(\cdot)$ is the Kronecker delta. This can be seen as no correlation between frequencies, which is the supposition of stationary. Moreover, the above equation is equivalent to the cross-spectral densities of the MOSM, and calculating the inverse generalised Fourier transform of these cross-spectral densities we obtain:

\begin{align*}
    \hat{k}(x,x')&=\iint_{\R^n\times\R^n}e^{\iu (\omega^{\top}x-\omega^{\prime Top})}\hat{S}_{ij}(\omega,\omega')d\omega d\omega'\\&= \iint_{\R^n\times\R^n}e^{\iu (\omega^{\top}x-\omega^{\prime Top})}\delta(\omega-\omega')e^{\left(-\frac{1}{2}(\overline{\omega}-\mu_{ij})^{\top}\Sigma_{ij}^{-1} (\overline{\omega}-\mu_{ij})\right)}e^{(-\iu(\theta_{ij}^{\top}\overline{\omega}+\phi_{ij}))}d\omega d\omega'\\
    &=\int_{\R^n}e^{\iu\omega^{\top}(x-x')}e^{\left(-\frac{1}{2}(\omega-\mu_{ij})^{\top}\Sigma_{ij}^{-1} (\omega-\mu_{ij})\right)}e^{(-\iu(\theta_{ij}^{\top}\omega+\phi_{ij}))}d\omega\\
    &=\alpha_{ij}e^{\left(-\frac{1}{2}(\tau+\theta_{ij})^{\top}\Sigma_{ij}(\tau+\theta_{ij})\right)}e^{\left(\iu(\tau+\theta_{ij})^{\top}\mu_{ij}+\phi_{ij}\right)}.
\end{align*}

Taking the real part of the above expression we recover the MOSM kernel. Thus, the MOHSM kernel successfully extend the MOSM kernel.

\section{EXPERIMENT DETAILS AND ADDITIONAL FIGURES}

\subsection{Learning derivatives and delayed signals}

we demonstrate the expressiveness of the MOHSM by using it to recover the auto and cross covariance of a MOGP. We considered an MOGP with the following three components: a sample $f$ from a GP with a non stationary kernel and zero mean, its derivative and a delayed version of the GP. This experiment is very illustrative since the covariance and cross covariance of the mentioned process are known explicitly, thus we can test the expressiveness of the model, namely

\begin{prop}
     Let $f$ be a Gaussian process with covariance function $k$, then the derivative stochastic process $f'$ is also a Gaussian process and its covariance function is $\frac{\partial^2 k(x,x')}{\partial x \partial x'}$. Furthermore, $(f(x),f'(x))$ form a two-channel Multi-Output Gaussian process \cite{williams2006gaussian} with the following multivariate covariance function
     \begin{equation}
     \mathcal{K}(x,x') = 
        \begin{pmatrix}
             k(x,x') & \frac{\partial k(x,x')}{\partial x'} \\
             \frac{\partial k(x,x')}{\partial x} & \frac{\partial^2 k(x,x')}{\partial x \partial x'}
        \end{pmatrix}
     \end{equation}
\end{prop}

\newpage

\subsection{GONU dataset}

In this section we present extra figures for the GONU experiment. We show the performance of each method with which we compare the MOHSM. 

\begin{figure}[h!]
    \centering
    \includegraphics[scale=0.3]{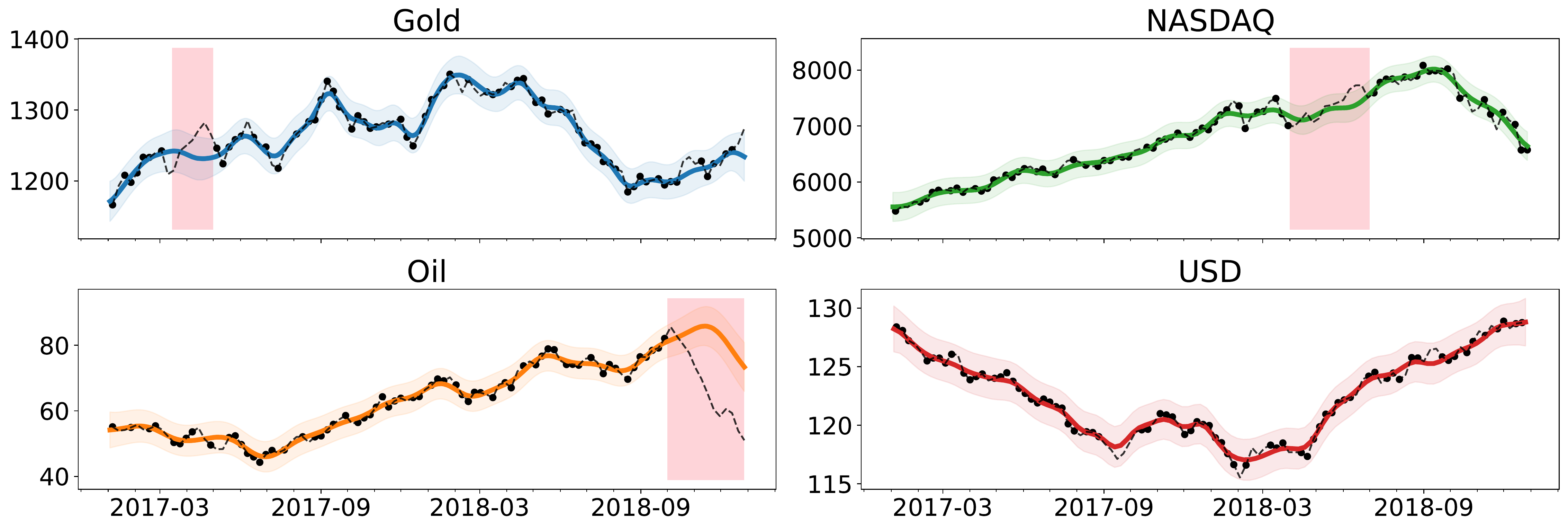}
    \caption{GONU data set with the trained MOSM kernel. Training points are shown in black, dashed lines are the ground truth and the colour coded lines are the posterior means. The coloured bands show the 95\% confidence intervals. The red shaded areas mark the data imputation ranges}
\end{figure}
\begin{figure}[h!]
    \centering
    \includegraphics[scale=0.3]{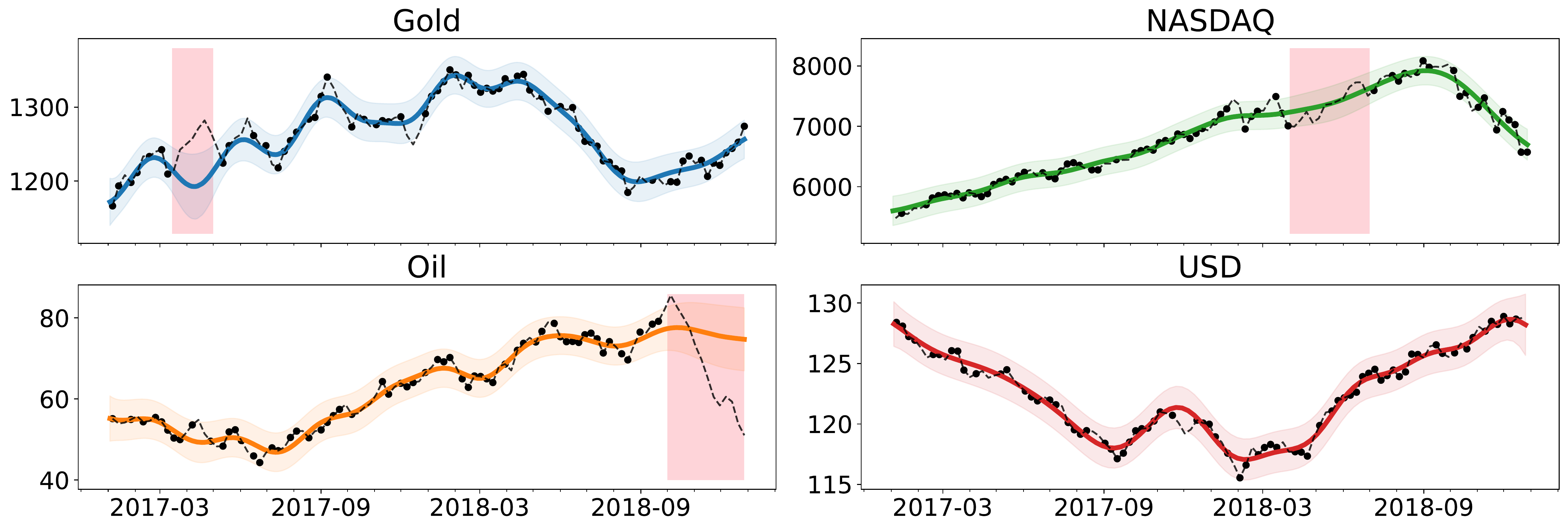}l
    \caption{GONU data set with the trained HSM-LMC kernel. Training points are shown in black, dashed lines are the ground truth and the colour coded lines are the posterior means. The coloured bands show the 95\% confidence intervals. The red shaded areas mark the data imputation ranges}
\end{figure}
\begin{figure}[h!]
    \centering
    \includegraphics[scale=0.3]{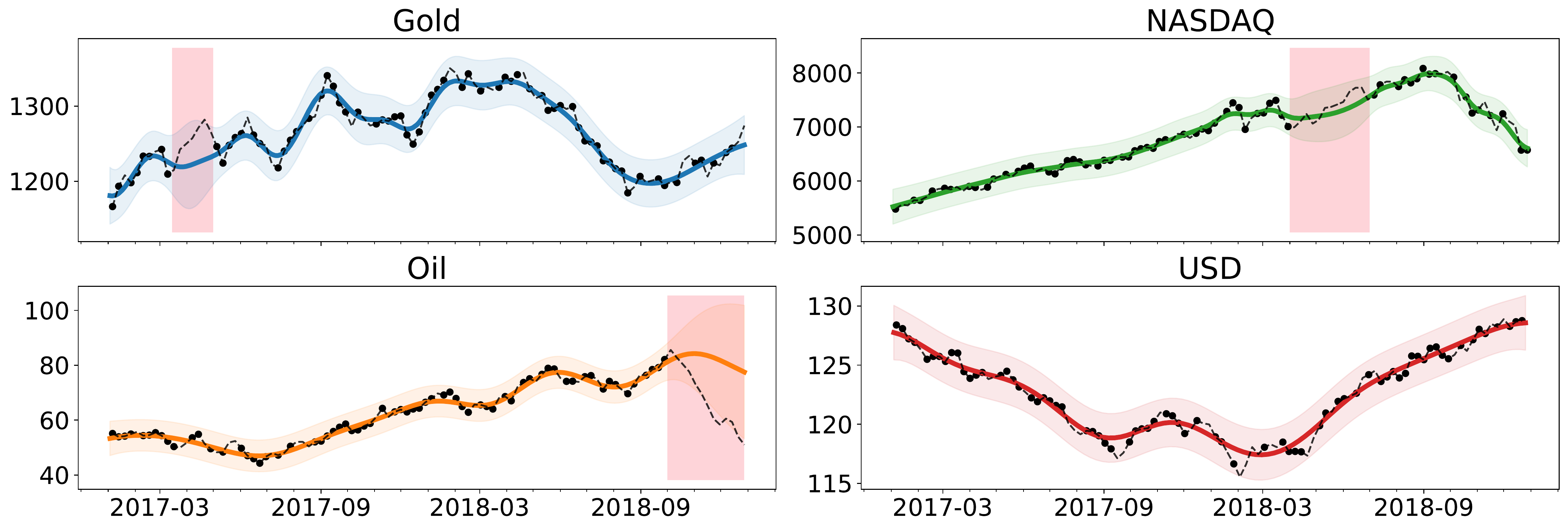}
    \caption{GONU data set with the trained HSM kernel. Training points are shown in black, dashed lines are the ground truth and the colour coded lines are the posterior means. The coloured bands show the 95\% confidence intervals. The red shaded areas mark the data imputation ranges}
\end{figure}

\newpage

\subsection{EEG dataset}

 \begin{figure}[h!]
      \centering
      \includegraphics[scale=0.25]{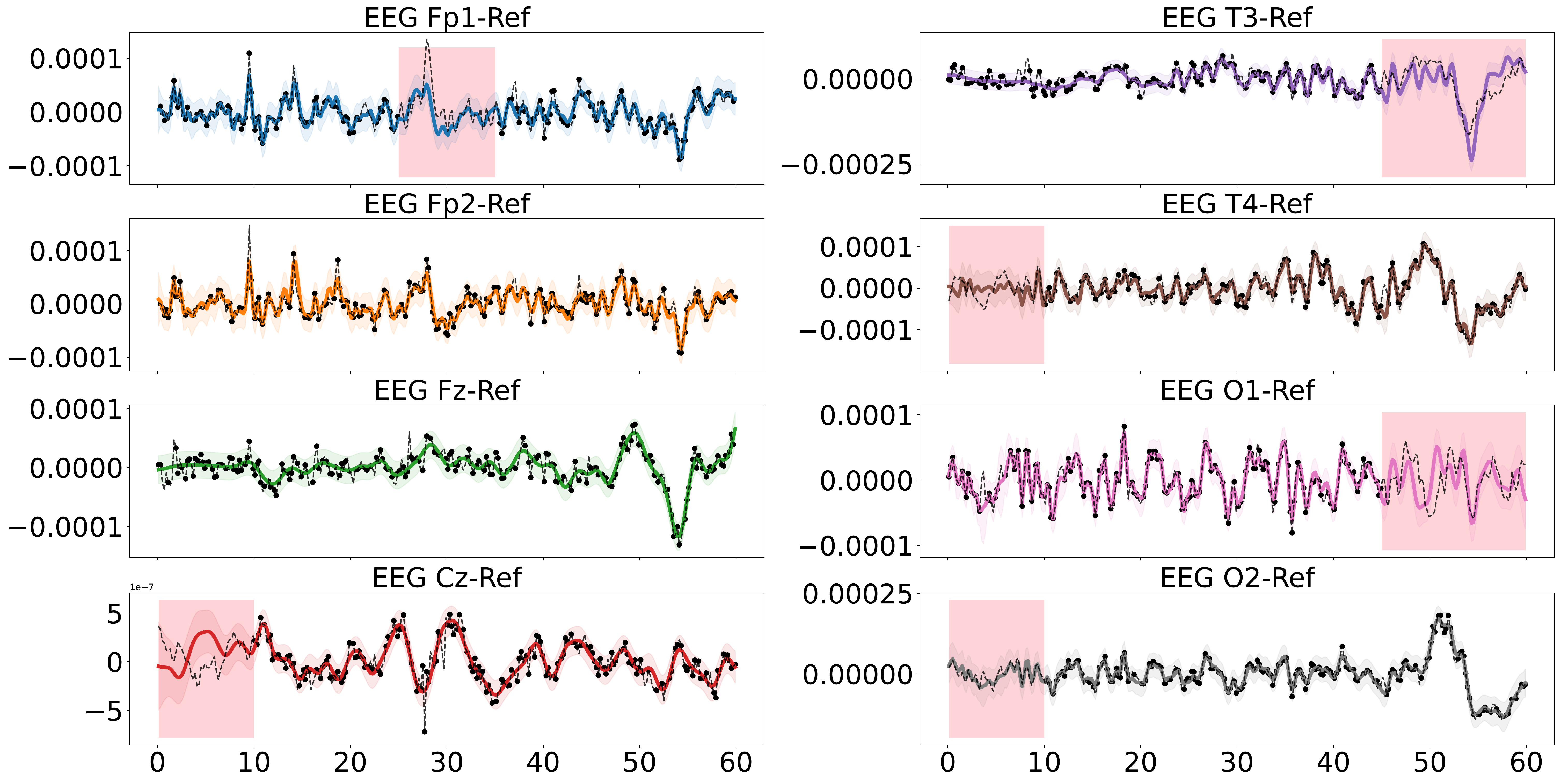}
      \caption{Extrapolation example using \textbf{MOHSM} kernel on EEG: training points (black), ground truth (dashed lines), and posterior means with 95\% error bars (color). The red shades denote data imputation ranges.}
\end{figure}
\end{document}